\documentclass{article}

\usepackage{arxiv}
\usepackage[english]{babel}
\usepackage{longtable}
\usepackage{array}
\usepackage{caption} % Optional, for more control over captions
\usepackage{booktabs} % For better table lines

\usepackage[utf8]{inputenc} % allow utf-8 input
\usepackage[T1]{fontenc}    % use 8-bit T1 fonts
\usepackage{hyperref}       % hyperlinks
\usepackage{url}            % simple URL typesetting
\usepackage{booktabs}       % professional-quality tables
\usepackage{amsfonts}       % blackboard math symbols
\usepackage{nicefrac}       % compact symbols for 1/2, etc.
\usepackage{microtype}      % microtypography
\usepackage{lipsum}
\usepackage{graphicx}
\usepackage{float}
\usepackage{amsmath,amssymb}
\usepackage{iftex}
\ifPDFTeX
  \usepackage[textcomp]{}
\else % if luatex or xetex
  \usepackage{unicode-math} % this also loads fontspec
  \defaultfontfeatures{Scale=MatchLowercase}
  \defaultfontfeatures[\rmfamily]{Ligatures=TeX,Scale=1}
\fi
\usepackage{lmodern}
\usepackage{xcolor}
\usepackage{longtable,booktabs,array}
\usepackage{calc} % for calculating minipage widths
\usepackage{etoolbox}

\makeatletter
\patchcmd\longtable{\par}{\if@noskipsec\mbox{}\fi\par}{}{}
\makeatother
\IfFileExists{footnotehyper.sty}{\usepackage{footnotehyper}}{\usepackage{footnote}}
\makesavenoteenv{longtable}

\newsavebox\pandoc
\newcommand*\pandocbounded[1]{% scales image to fit in text height/width
  \sbox\pandoc@box{#1}%
  \Gscale@div\@tempa{\textheight}{\dimexpr\ht\pandoc@box+\dp\pandoc@box\relax}%
  \Gscale@div\@tempb{\linewidth}{\wd\pandoc@box}%
  \ifdim\@tempb\p@<\@tempa\p@\let\@tempa\@tempb\fi% select the smaller of both
  \ifdim\@tempa\p@<\p@\scalebox{\@tempa}{\usebox\pandoc@box}%
  \else\usebox{\pandoc@box}%
  \fi%
}
\title{Leveraging Natural Language and Item Response Theory Models for ESG Scoring}

\author{
 César P. Soares \\
  Ph.D. in Global Health and Sustainability \\
  \texttt{cpscesar@gmail.com } \\
}

\begin{document}
\maketitle

\begin{abstract}
This paper explores an innovative approach to Environmental, Social, and Governance (ESG) scoring by integrating Natural Language Processing (NLP) techniques with Item Response Theory (IRT), specifically the Rasch model. The study utilizes a comprehensive dataset of news articles in Portuguese related to Petrobras, a major oil company in Brazil, collected from 2022 and 2023. The data is filtered and classified for ESG-related sentiments using advanced NLP methods. The Rasch model is then applied to evaluate the psychometric properties of these ESG measures, providing a nuanced assessment of ESG sentiment trends over time. The results demonstrate the efficacy of this methodology in offering a more precise and reliable measurement of ESG factors, highlighting significant periods and trends. This approach may enhance the robustness of ESG metrics and contribute to the broader field of sustainability and finance by offering a deeper understanding of the temporal dynamics in ESG reporting.
\end{abstract}


\section{Introduction}
In recent years, the integration of language models into various facets of research and industry has ushered in a new era of text generation and comprehension. These models, driven by advancements in deep learning and natural language processing (NLP), hold the potential to change how we understand and interact with textual data.

The convergence of NLP and psychometric methods are emerging in this context as an interdisciplinary approach for tackling complex socio-environmental analyses \cite{lalor2016building, johannssen2018lines, kennedy2020constructing, rathi2022psychometric, soares2022proposal}. By combining both, researchers aim to unlock new insights into the intricate interplay between human behaviour and nature aspects. This paper leverages this interdisciplinary synergy to propose a new methodology to help estimate Environmental, Social, and Governance (ESG) rating scores. Integrating psychometric techniques allows for the rigorous measurement of latent constructs underlying ESG performance, while NLP techniques facilitate the extraction of valuable insights from vast troves of unstructured textual data.

ESG factors have emerged as critical for investors, stakeholders, and regulators seeking to evaluate companies and entities' sustainability and societal impact. Integrating ESG criteria into investment decisions reflects a growing recognition of the interconnectedness between corporate performance, environmental stewardship, social responsibility, and effective governance practices. Consequently, there is an increasing demand for robust methodologies and tools to assess and quantify ESG performance accurately \cite{zumente2021esg,michalski2021corporate,zairis2024sustainable}.

Previous studies in the realm of ESG scoring have explored the application of NLP techniques to analyze textual data extracted from various sources such as news articles, social media, and corporate reports \cite{sokolov2020building,aue2022predicting,atanassova2024criticalminds}. However, notably absent from these endeavours is the integration of psychometric methods, which may work instrumentally in assessing the reliability and validity of the measured ESG construct. While NLP offers unparalleled capabilities in processing and extracting insights from unstructured textual data, the absence of psychometric validation may introduce potential limitations in the robustness and interpretability of the resulting ESG ratings. By exploring the possibility of incorporating psychometric principles into the ESG measurement framework, our research aims to address this gap by providing a more comprehensive and rigorous approach to ESG scoring, thereby enhancing the accountability and reliability of the assessment process.

In other words, this paper aims to explore a novel methodology that integrates NLP and Item Response Theory (IRT) to enhance the robustness of studies calculating ESG metrics. While the paper does not propose a specific metric, the combined NLP and IRT results presented may contribute to further developing more reliable and robust ESG metric calculations. The results may also contribute to advancing scholarship at the intersection of finance, sustainability, and computational linguistics by shedding light on the transformative potential of labelled text data for enhancing ESG estimation methodologies.

This paper is organized as follows: Section 2 presents the methodology. Section 3 presents the results. Section 4 discusses the implications of our findings and potential applications. Finally, Section 5 concludes the paper and outlines future research directions.

\section{Methodology}

\subsection{Study Design}

This study's methodology leverages labelled text data to enhance the estimation of the ESG construct in a similar way applied by Aue, Jatowt \& Färber \cite{aue2022predicting}. Labelled text data consists of textual information annotated according to specific ESG-related dimensions, including environmental impact, social responsibility, and corporate governance practices. These data sources encompass news articles, offering valuable insights for ESG analysis. Advanced NLP techniques are employed to extract meaningful information from these textual sources.

Incorporating IRT into the methodology provides a robust framework for assessing the psychometric properties of ESG constructs. IRT facilitates the measurement of latent traits by analyzing the relationship between news data and the underlying trait being measured. This study uses the Rasch model to evaluate the consistency and reliability of ESG measures derived from textual data.

The first step involved downloading and scraping news articles in Portuguese using the Global Database of Events, Language, and Tone (GDEL)\footnote{More information: \url{https://www.gdeltproject.org/}}. The focus was on Brazil in 2022 and 2023, particularly the company Petrobras\footnote{More information: \url{https://petrobras.com.br/en/}}, a significant oil business in Brazil. A comprehensive dataset covering these years was organized by concatenating data from all months. Each month was treated as an individual "item" to be lately evaluated by the Rasch model.

The second step involved cleaning the dataset using Bidirectional Encoder Representations from Transformers (BERT) for the Portuguese language to apply embeddings and similarity algorithms with ESG definitions. This process filtered relevant and irrelevant news articles, retaining only those pertinent to one of the ESG dimensions. The final dataset considered 3,653 news for 2022 and 4,401 news for 2023.

In the third step, the dataset was processed using embeddings and similarity algorithms to classify a sample of 365 cases, or 10\% of 2022 news articles' data, as positive or negative. A Portuguese BERT classification model, trained and fine-tuned on this data sample, determined the sentiment of news articles regarding ESG dimensions on the whole dataset of 2022 and 2023. The table below shows the ESG definitions and the positive/negative criteria used for similarity analysis. The text is in Portuguese since the news texts were in this language.

% Ensure the table title and table are on the same page
\captionsetup{justification=centering, singlelinecheck=false}

\begin{table}[H]
\centering
\caption{ESG Definitions}
\begin{longtable}{|>{\centering\arraybackslash}m{0.15\textwidth}|
                    >{\centering\arraybackslash}m{0.3\textwidth}|
                    >{\centering\arraybackslash}m{0.25\textwidth}|
                    >{\centering\arraybackslash}m{0.25\textwidth}|}
\hline
\textbf{ESG Dimension} & \textbf{ESG Definition} & \textbf{Positive ESG} & \textbf{Negative ESG} \\
\hline
\endfirsthead
\hline
\textbf{ESG Dimension} & \textbf{ESG Definition} & \textbf{Positive ESG} & \textbf{Negative ESG} \\
\hline
\endhead
\hline
\textbf{Environment} & São reportados dados sobre alterações climáticas, emissões de gases com efeito estufa, perda de biodiversidade, desflorestação/reflorestação, mitigação da poluição, eficiência energética e gestão da água. & Iniciativas para a mitigação das alterações climáticas, redução das emissões de gases com efeito de estufa, promoção da biodiversidade, reflorestação, redução da poluição, melhoria da eficiência energética e gestão eficaz da água. & Questões relacionadas com a inação em matéria de alterações climáticas, o aumento das emissões de gases com efeito de estufa, a perda de biodiversidade, a desflorestação, a poluição, a utilização ineficiente da energia e a má gestão da água. \\
\hline
\textbf{Social} & Os dados são relatados sobre segurança e saúde dos funcionários, condições de trabalho, diversidade, equidade e inclusão, e conflitos e crises humanitárias, e são relevantes nas avaliações de risco e retorno diretamente através dos resultados no aumento (ou destruição) da satisfação do cliente e da satisfação dos funcionários. & Melhorias na saúde e segurança dos funcionários, melhores condições de trabalho, avanços na diversidade e inclusão, resolução de conflitos, maior satisfação do cliente e envolvimento dos funcionários. & Problemas com a saúde e segurança dos funcionários, más condições de trabalho, falta de diversidade e inclusão, conflitos não resolvidos, baixa satisfação dos clientes e envolvimento dos funcionários. \\
\hline
\textbf{Governance} & Os dados são relatados sobre governança corporativa, como prevenção de suborno, corrupção, diversidade do conselho de administração, remuneração de executivos, práticas de segurança cibernética, privacidade e estrutura de gestão. & Governança corporativa forte, medidas antissuborno, diversidade do conselho, remuneração justa dos executivos, práticas robustas de segurança cibernética e privacidade. & Governança corporativa fraca, problemas com suborno e corrupção, falta de diversidade no conselho, remuneração inadequada dos executivos, práticas inadequadas de segurança cibernética e privacidade. \\
\hline
\end{longtable}
\vspace{-60pt} % Adjust this value as needed
\end{table}

\noindent\textbf{Source:} Built by the author, 2024, based on Wikipedia and OpenAI ChatGPT consultation.

\vspace{10pt}
The final step involved structuring the dichotomous dataset, where each
row represented the sentiment of the news articles (1 for positive and 0
for negative), and each column represented a month of the year 2022 and
2023. Then, we run the Rasch psychometric model to obtain the
psychometric characteristics of the ESG construct.

\subsection{Training Dataset}

We used a sample of news from 2022 (n = 365) to build the dataset for
training the model. All the news was in Portuguese.

The training and evaluation pipeline used a pre-trained BERTimbau Base
model for feature extraction and a custom classifier for classification.
The training was done in two stages: the first stage only trained the
classifier, and the second stage fine-tuned the entire model.

The results obtained by the training process were logged in a CSV file,
including various metrics that were further analyzed to determine the
best model. The steps for this procedure are below.

\begin{figure}[H]
    \centering
    \includegraphics[width=0.2\textwidth]{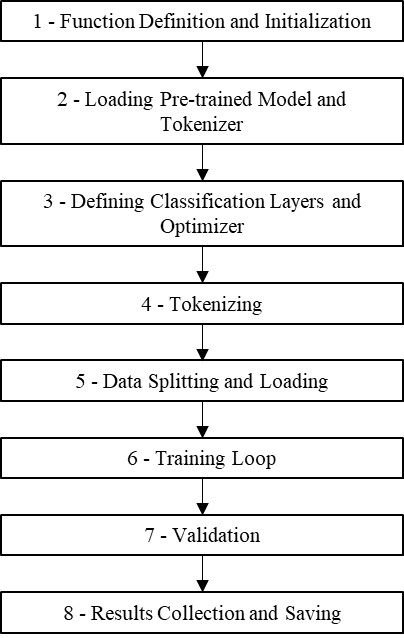} % Adjust the width as needed
    \vspace{10pt} % Adjust vertical space between image and caption
    \caption{Steps flow of training}
    \label{fig:training_flow}
    \vspace{5pt} % Space between caption and source
    \textbf{Source:} Built by the author, 2024.
\end{figure}

% Space between the figure and the following text
\vspace{20pt}

% Start new section after the figure
\begin{enumerate}
\def\labelenumi{\arabic{enumi}.}
\item
  \emph{Function Definition and Initialization}
\end{enumerate}

The code first defined a function that took several parameters (e.g.,
learning rate, batch size, epochs). Below, you can see the combination
of tested parameters. In the end, we tested 729 combinations of
parameter values. The last subsection of the methodology explains how we
analyzed the metrics to define the best model to be used.
\vspace{20pt}

\textbf{Table 2:} Parameters considered

\begin{longtable}[]{@{}
  >{\centering\arraybackslash}p{(\linewidth - 6\tabcolsep) * \real{0.2499}}
  >{\centering\arraybackslash}p{(\linewidth - 6\tabcolsep) * \real{0.2499}}
  >{\centering\arraybackslash}p{(\linewidth - 6\tabcolsep) * \real{0.2501}}
  >{\centering\arraybackslash}p{(\linewidth - 6\tabcolsep) * \real{0.2501}}@{}}
\toprule\noalign{}
\begin{minipage}[b]{\linewidth}\centering
\textbf{Parameters}
\end{minipage} &
\multicolumn{3}{>{\centering\arraybackslash}p{(\linewidth - 6\tabcolsep) * \real{0.7501} + 4\tabcolsep}@{}}{%
\begin{minipage}[b]{\linewidth}\centering
\textbf{Values}
\end{minipage}} \\
\midrule\noalign{}
\endhead
\bottomrule\noalign{}
\endlastfoot
Learning rate & 1e-5 & 2e-5 & 3e-5 \\
Number of layers & 2 & 5 & 10 \\
Hidden Size & 256 & 512 & 768 \\
Batch Size & 5 & 10 & 20 \\
Epochs & 2 & 5 & 10 \\
Max Length & 50 & 100 & 200 \\

\end{longtable}
\vspace{-10pt} % Adjust this value as needed

\textbf{Source:} build by the author, 2024.
\vspace{20pt}

\begin{enumerate}
\def\labelenumi{\arabic{enumi}.}
\setcounter{enumi}{1}
\item
  \emph{Loading Pre-trained Model and Tokenizer}
\end{enumerate}

Based on the function definition above, the initialization started
loading the pre-trained BERTimbau Base model and tokenizer from the
Hugging Face Transformers library. We tokenized the input text (news)
and converted it into input IDs and attention masks. These tokenized
sequences were fed into the model, which produced high-dimensional
contextualized embeddings for each token in the sequence. These
embeddings captured the semantic information of the words in the context
of the entire sequence.

\begin{enumerate}
\def\labelenumi{\arabic{enumi}.}
\setcounter{enumi}{2}
\item
  \emph{Defining Classification Layers and Optimizer}
\end{enumerate}

A custom classification layer was defined using
PyTorch's `Sequential` module. It consisted of a Long
Short-Term Memory (LSTM) layer followed by a dropout layer to prevent
overfitting and a linear layer to map the LSTM's outputs
to the final classification decision. The softmax activation function
was applied to the output of this linear layer to produce class
probabilities for news sentiment classification analysis. This
architecture allowed the model to learn temporal dependencies and
patterns in the sequence of embeddings.

The purpose of this classifier was to learn patterns and relationships
in the embeddings that were relevant to the classification task.
Specifically, the custom classifier was a separate neural network built
on the embedding model. It was designed to take the contextualized
embeddings and make a final classification decision (e.g., positive or
negative in the case of the news considered in this paper). We
initialized this classifier with random weights and trained it to fit
the classification task.

\begin{enumerate}
\def\labelenumi{\arabic{enumi}.}
\setcounter{enumi}{3}
\item
  \emph{Tokenizing}
\end{enumerate}

The news text was tokenized using the previously loaded tokenizer. It
then converted the tokenized data into input IDs and attention masks.
The input IDs, attention masks, and labels were stored in separate lists
for use in the next steps.

\begin{enumerate}
\def\labelenumi{\arabic{enumi}.}
\setcounter{enumi}{4}
\item
  \emph{Data Splitting and Loading}
\end{enumerate}

The data was divided into training (n = 329, 90\%) and validation (n =
36, 10\%) sets. We defined data loaders for batch-wise processing.

\begin{enumerate}
\def\labelenumi{\arabic{enumi}.}
\setcounter{enumi}{5}
\item
  \emph{Training Loop}
\end{enumerate}

\begin{quote}
The code had a two-stage training loop. The first stage involved
freezing the model and training only the custom classifier. The second
stage unfreezes the model for fine-tuning:
\end{quote}

\begin{itemize}
\item
  \textbf{Stage 1 (Classifier Training):}

  \begin{itemize}
  \item
    \textbf{Objective:} In this initial stage, the primary goal was to
    adapt a pre-trained model for our specific text classification task.
    This process aimed to "teach" the model to classify text into
    different categories or labels (e.g. positive or negative).
  \item
    \textbf{Model Configuration:} At the beginning of Stage 1, the model
    was set to "evaluation mode," and we froze its parameters. This
    process means that we did not update the model during this stage. It
    remained as it was after pre-training on a large corpus.
  \item
    \textbf{Custom Classifier:} In addition to the model, we had a
    custom classifier, which typically consists of one or more layers
    (in this case, an LSTM layer and a linear layer). These layers were
    responsible for taking the representations learned by the model and
    making the final classification decision.
  \item
    \textbf{Training Focus:} During Stage 1, the only updated parameters
    were those of the custom classifier (LSTM and linear layers). The
    model's parameters remained fixed. This stage aimed
    to fine-tune the custom classifier's parameters to
    be well-suited for the specific classification task. It taught the
    model how to use its pre-trained embeddings to classify text.
  \end{itemize}
\item
  \textbf{Stage 2 (Fine-tuning):}

  \begin{itemize}
  \item
    \textbf{Objective:} After completing Stage 1, the model had a custom
    classifier that was better adapted to the classification task.
    However, the pre-trained model still had generic language
    representations. In Stage 2, the goal was to further adapt the
    entire model for the specific classification task.
  \item
    \textbf{Model Configuration:} Unlike Stage 1, this stage, it was
    allowed updates to both the custom classifier (LSTM and linear
    layers) and the model.
  \item
    \textbf{Training Focus:} During Stage 2, the model was fine-tuned
    using the labelled data for the classification task. This process
    allowed the model to adjust the learned representations of the text
    data, including those learned before, to fit the classification
    task's nuances better. Essentially, it allowed the
    model to make more significant changes to its internal
    representations based on the specific data it was working with.
  \end{itemize}
\end{itemize}

This two-stage process was designed to leverage the power of pre-trained
language models like BERTimbau Base while still tailoring them to this
paper's specific NLP classification task. Stage 1
focused on adapting the classifier, and Stage 2 further adapted the
entire model to maximize its performance on the task at hand.

\begin{enumerate}
\def\labelenumi{\arabic{enumi}.}
\setcounter{enumi}{6}
\item
  \emph{Validation}
\end{enumerate}

After each epoch, the classifier was set to evaluation mode, iterating
over the validation data and calculating loss and accuracy.
Additionally, it computed precision, recall, F1-score, and AUC-ROC for
each class (e.g., positive and negative).

\begin{enumerate}
\def\labelenumi{\arabic{enumi}.}
\setcounter{enumi}{7}
\item
  \emph{Results Collection and Saving}
\end{enumerate}

We computed metrics, and results were collected and written in the
output CSV file.

\subsection{Evaluating the model}

The trained classifier was evaluated using the validation data. We
passed the validation data through the model, and the logits were
obtained. The cross-entropy loss was computed, and evaluation metrics
such as accuracy, precision, recall, F1 score, and area under the ROC
curve (AUC) were calculated. We used the Technique for Order of
Preference by Similarity to Ideal Solution (TOPSIS) scores for ranking.

Based on the evaluation metrics calculated during the finetuning
process, we implemented a normalization function to normalize the values
using a weighted approach. For each value, the function calculated the
root of the sum of squares and then applied weighted normalization to
each element. The weights were determined based on the literature on the
area and the researcher's knowledge of the importance of
the variables.

Further, the score calculation involved computing the distance from each
column's ideal best and worst values. We summed up these
distances after squaring each value. Finally, the TOPSIS score for each
model was computed as the distance ratio from the ideal worst to the sum
of the distances from the ideal best and ideal worst. The ranking of the
best models was derived based on their TOPSIS scores.
\vspace{20pt}

\textbf{Table 3:} TOPSIS weights

\begin{longtable}[]{@{}
  >{\centering\arraybackslash}p{(\linewidth - 16\tabcolsep) * \real{0.0728}}
  >{\centering\arraybackslash}p{(\linewidth - 16\tabcolsep) * \real{0.1268}}
  >{\centering\arraybackslash}p{(\linewidth - 16\tabcolsep) * \real{0.0846}}
  >{\centering\arraybackslash}p{(\linewidth - 16\tabcolsep) * \real{0.1014}}
  >{\centering\arraybackslash}p{(\linewidth - 16\tabcolsep) * \real{0.1394}}
  >{\centering\arraybackslash}p{(\linewidth - 16\tabcolsep) * \real{0.1100}}
  >{\centering\arraybackslash}p{(\linewidth - 16\tabcolsep) * \real{0.0962}}
  >{\centering\arraybackslash}p{(\linewidth - 16\tabcolsep) * \real{0.1344}}
  >{\centering\arraybackslash}p{(\linewidth - 16\tabcolsep) * \real{0.1344}}@{}}
\toprule\noalign{}
\begin{minipage}[b]{\linewidth}\centering
\textbf{Train}

\textbf{loss}
\end{minipage} & \begin{minipage}[b]{\linewidth}\centering
\textbf{Train accuracy}
\end{minipage} & \begin{minipage}[b]{\linewidth}\centering
\textbf{Val.}

\textbf{loss}
\end{minipage} & \begin{minipage}[b]{\linewidth}\centering
\textbf{Val.}

\textbf{accuracy}
\end{minipage} & \begin{minipage}[b]{\linewidth}\centering
\textbf{Val. precision}
\end{minipage} & \begin{minipage}[b]{\linewidth}\centering
\textbf{Val. recall}
\end{minipage} & \begin{minipage}[b]{\linewidth}\centering
\textbf{Val.}

\textbf{F1 score}
\end{minipage} & \begin{minipage}[b]{\linewidth}\centering
\textbf{Val. auc roc1}
\end{minipage} & \begin{minipage}[b]{\linewidth}\centering
\textbf{Val. auc roc2}
\end{minipage} \\
\midrule\noalign{}
\endhead
\bottomrule\noalign{}
\endlastfoot
0.1 & 0.1 & 0.2 & 0.2 & 0.1 & 0.1 & 0.1 & 0.1 & 0.1 \\
\end{longtable}
\vspace{-10pt}
\textbf{Source:} build by the author, 2024. ``Val.'' means Validation.
``auc roc'' means Area Under the Receiver Operating Characteristic
Curve.

\vspace{20pt}
\subsection{IRT -- Rasch Model}

The final step involved structuring the dichotomous dataset, where each
row represented the sentiment of the news articles (1 for positive and 0
for negative), and each column represented a month of the years 2022 and
2023. This arrangement allowed for a comprehensive temporal analysis of
sentiment trends across the specified period.

IRT is a family of mathematical models used to analyze the relationship
between individuals' latent traits (unobservable
characteristics or attributes) and their item responses on assessments
or questionnaires. IRT provides a framework for understanding how
specific test items function across different latent trait levels. It is
often used in educational testing, psychological assessments, and health
outcomes measurement. IRT models include the Rasch model, the one and
two-parameter logistic model, and the three-parameter logistic model,
each differing in complexity and assumptions about the data \cite{bond2013rasch}.

We employed the Rasch psychometric model to analyze the psychometric
properties of the ESG construct. The Rasch model, a specific form of
IRT, is particularly suitable for dichotomous data and is widely used in
psychometrics. It facilitates the measurement of latent traits by
modelling the probability of a given response as a function of the
respondent's ability and the item's
difficulty. This model helped understand how different periods
(represented by months) contributed to the overall ESG sentiment,
enabling a detailed assessment of its psychometric characteristics.
Moreover, the "individual" being analyzed is the Petrobras company, and
the news articles served as the responses to the items in the dataset.
The primary objective was not to examine the specific result of each
news item but to assess how each month's positive or
negative sentiment affected the ESG construct.

This analysis helped us to understand the temporal dynamics of ESG
sentiment for Petrobras based on media news and may guide future
strategies to measure ESG. By structuring the dataset to reflect monthly
sentiment trends, we can gain valuable insights into how different
periods impact the company's ESG profile.

When applied to ESG measurement, the concepts of IRT might not directly
translate into the traditional sense used in educational or
psychological assessments. Instead of "proficiency," the latent trait
measured is the sentiment level regarding ESG news. This sentiment level
indicates how positive or negative the news is perceived. Similarly,
item difficulty (monthly difficulty parameters) is understood
differently. Higher difficulty values for a particular month indicate
that achieving positive news during that month was harder. This context
could be due to more negative news or challenging ESG-related events.

To analyze these factors, we presented each month's Item
Information Curves (IIC) and Item Characteristic Curves (ICC). The IICs
show how much information each month provides about the sentiment
levels. Peaks at different sentiment levels indicate that certain months
are more informative for different sentiment ranges. The ICC illustrates
the probability of receiving a positive/negative sentiment score as a
function of the underlying sentiment trait.

We can understand the temporal trends in ESG sentiment by analyzing the
difficulty parameters and information curves. Months with greater
difficulty and more information could indicate critical periods where
significant ESG events occurred. These periods are particularly
informative for understanding the challenges and achievements in the
company's ESG initiatives. This comprehensive analysis
highlights key periods of ESG performance and provides a framework for
ongoing monitoring and improvement of ESG strategies.

\section{Results}

Considering the TOPSIS score, the model below presented the best
results:
\vspace{10pt}

\textbf{Table 4:} Best model parameters combination
\vspace{-10pt}
\begin{longtable}[]{@{}
  >{\centering\arraybackslash}p{(\linewidth - 2\tabcolsep) * \real{0.5786}}
  >{\centering\arraybackslash}p{(\linewidth - 2\tabcolsep) * \real{0.4214}}@{}}
\toprule\noalign{}
\begin{minipage}[b]{\linewidth}\centering
\textbf{Parameters}
\end{minipage} & \begin{minipage}[b]{\linewidth}\centering
\textbf{Best Model}
\end{minipage} \\
\midrule\noalign{}
\endhead
\bottomrule\noalign{}
\endlastfoot
Learning rate & 2e-5 \\
Number of layers & 5 \\
Hidden Size & 768 \\
Batch Size & 20 \\
Epochs & 10 \\
Max Length & 200 \\
\end{longtable}
\vspace{-20pt}
\textbf{Source:} build by the author, 2024.

The confusion matrix below shows how the model classified the news after the training.\\

\textbf{Fig. 2:} Confusion matrix

\includegraphics[width=3.91135in,height=3.01325in]{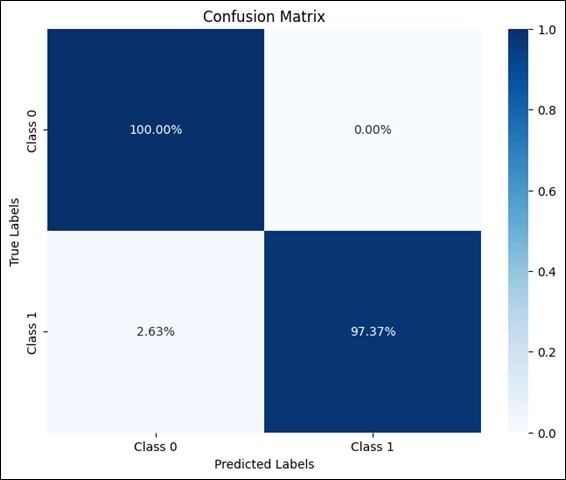}

\textbf{Source:} build by the author, 2024. Class 0: Negative ESG News.
Class 1: Positive ESG News.

The confusion matrix results indicate that the classification model used
to distinguish between negative and positive ESG news is highly
effective. The model correctly identified 100\% of the negative ESG news
(Class 0) without any false positives, demonstrating precision for
negative news. It also achieved a 97.37\% accuracy in identifying
positive ESG news (Class 1), with a false negative rate of 2.63\%.

The figures below show the IIC for 2022 and 2023.
\vspace{10pt}

\textbf{Fig. 3:} IIC, 2022

\begin{minipage}[b]{0.48\textwidth}
    \centering
    \includegraphics[width=\textwidth]{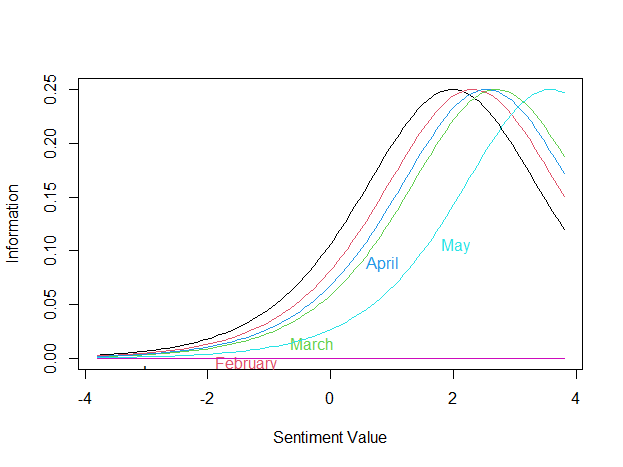}
    %\caption*{(a)} % Uncomment this line to add a sub-caption if needed
\end{minipage}
\hfill
\begin{minipage}[b]{0.48\textwidth}
    \centering
    \includegraphics[width=\textwidth]{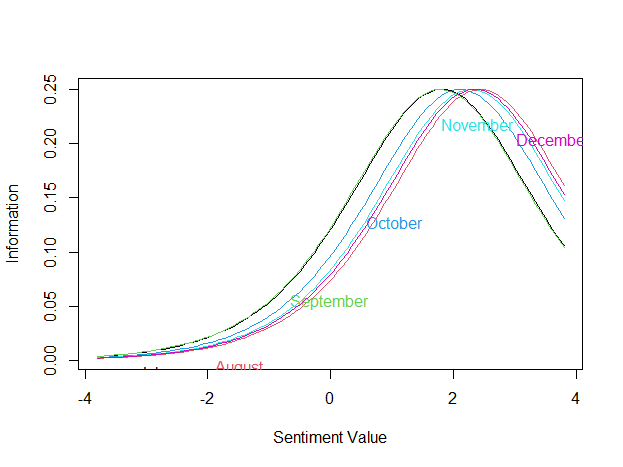}
    %\caption*{(b)} % Uncomment this line to add a sub-caption if needed
\end{minipage}

\textbf{Source:} build by the author, 2024.

\vspace{10pt}
\textbf{Fig. 4:} IIC, 2023

\begin{minipage}[b]{0.48\textwidth}
    \centering
    \includegraphics[width=\textwidth]{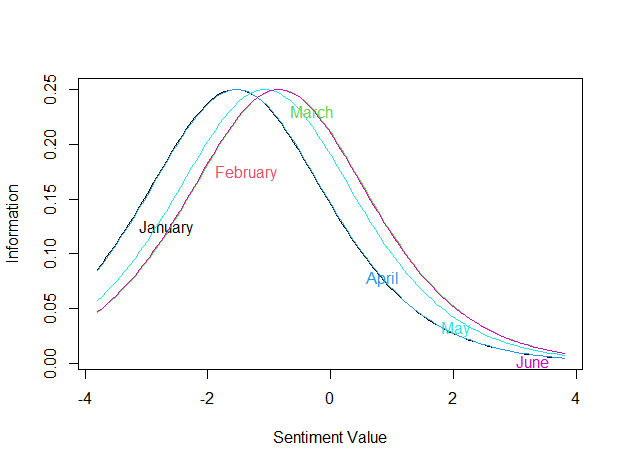}
    %\caption*{(a)} % Uncomment this line to add a sub-caption if needed
\end{minipage}
\hfill
\begin{minipage}[b]{0.48\textwidth}
    \centering
    \includegraphics[width=\textwidth]{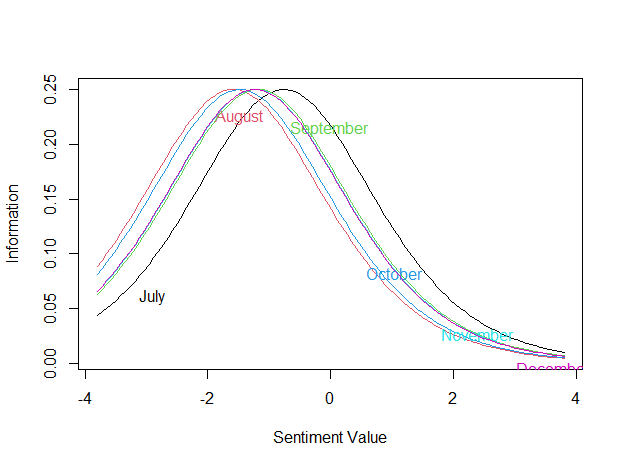}
    %\caption*{(b)} % Uncomment this line to add a sub-caption if needed
\end{minipage}

\textbf{Source:} build by the author, 2024.
\vspace{10pt}

Each curve on the IIC shows the amount of information a particular month
provides across these sentiment difficulty levels. Peaks at different
points indicate which months are more informative for different ranges
of sentiment difficulty. For instance, months with peaks at higher
sentiment values are more informative for capturing difficult (negative)
sentiment, while those peaking at lower values are more informative for
easier (positive) sentiment.

The figures below show the ICC for 2022 and 2023.
\vspace{100pt}

\textbf{Fig. 5:} ICC, 2022

\begin{minipage}[b]{0.48\textwidth}
    \centering
    \includegraphics[width=\textwidth]{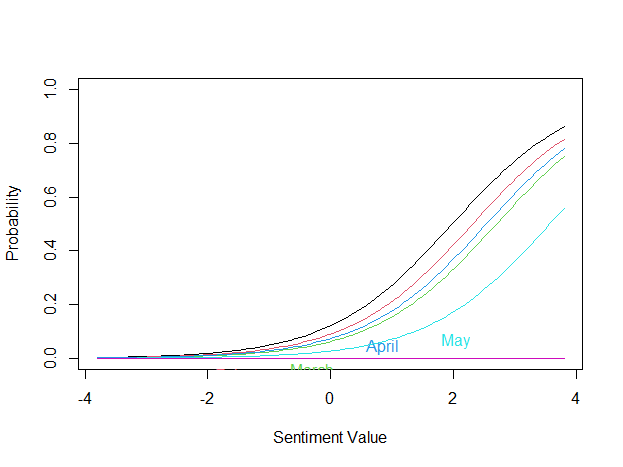}
    %\caption*{(a)} % Uncomment this line to add a sub-caption if needed
\end{minipage}
\hfill
\begin{minipage}[b]{0.48\textwidth}
    \centering
    \includegraphics[width=\textwidth]{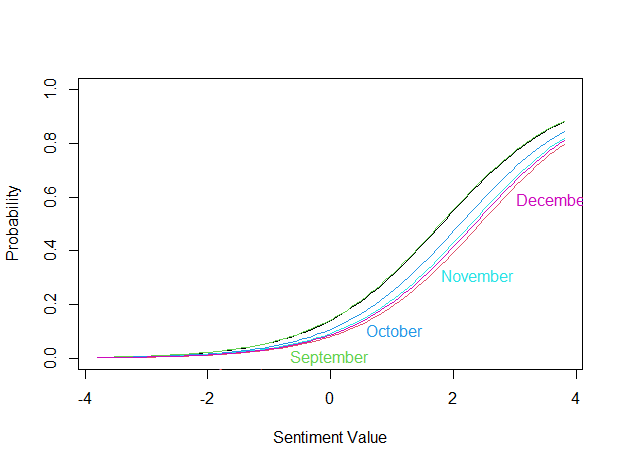}
    %\caption*{(b)} % Uncomment this line to add a sub-caption if needed
\end{minipage}

\textbf{Source:} build by the author, 2024.
\vspace{10pt}

\textbf{Fig. 6:} ICC, 2023

\begin{minipage}[b]{0.48\textwidth}
    \centering
    \includegraphics[width=\textwidth]{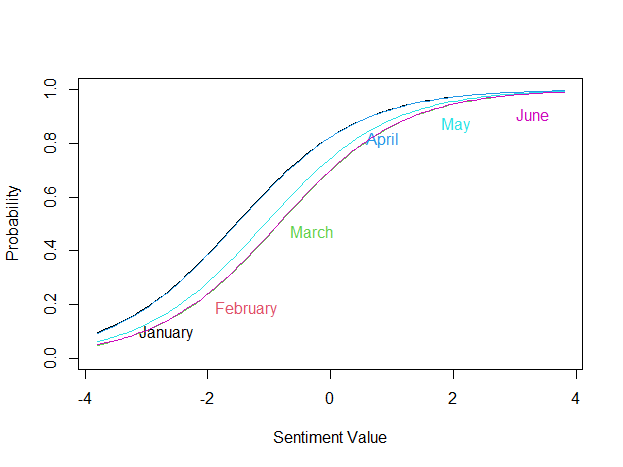}
    %\caption*{(a)} % Uncomment this line to add a sub-caption if needed
\end{minipage}
\hfill
\begin{minipage}[b]{0.48\textwidth}
    \centering
    \includegraphics[width=\textwidth]{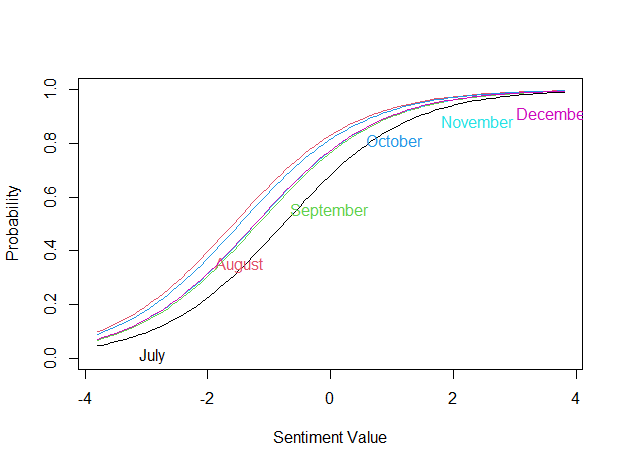}
    %\caption*{(b)} % Uncomment this line to add a sub-caption if needed
\end{minipage}

\textbf{Source:} build by the author, 2024.
\vspace{10pt}

The ICCs for 2022 and 2023 illustrate the probability of achieving
positive ESG sentiment across different months. Considering that a
sentiment value of 1 is positive, we move from left to right on the
x-axis from easiest (most positive news) to most difficult (most
negative news). These curves help identify key periods with significant
positive or negative ESG events, aiding in understanding temporal
dynamics in ESG sentiment and guiding strategic ESG measurement
improvements.

\section{Discussion}

This study has explored the potential of using IRT, specifically the
Rasch model, to analyze ESG sentiment data. Applying the Rasch model in
this context may enhance the measurement methods of this construct.

One of the main advantages of using the Rasch model is its ability to
provide a more precise measurement of ESG sentiment. By accounting for
both item difficulty (in this case, variability in sentiment across
different months) and respondent ability (in this case, the Brazillian
company Petrobras), the model places all data on an interval scale. This
not only facilitates more accurate comparisons over time but also
ensures that these comparisons are independent of the specific sample
used \cite{bond2013rasch}. The scale independence may be important for
ESG scoring process, where variability can arise from numerous external
factors, such as market events or regulatory changes.

The Rasch model's capacity to handle variability in item
characteristics allows for a nuanced understanding of how different
months contribute to the overall ESG sentiment. This is particularly
valuable for organizations aiming to track sentiment trends related to
ESG topics, as it can identify periods of heightened scrutiny or
significant sentiment shifts.

The use of IICs and ICCs in the Rasch model provides additional insight
into the informativeness of each month. This helps in identifying which
time periods offer the most significant data for assessing ESG
sentiment, thereby guiding more targeted analyses and strategic
responses. Additionally, the latent trait estimation offered by the
Rasch model may contribute to a more accurate depiction of the
underlying sentiment trends than simple averages, which can be skewed by
outliers or non-normal data distributions, or techniques used on the
studies of the area \cite{sokolov2020building,aue2022predicting, atanassova2024criticalminds}.

The precision and reliability of the Rasch model may contribute to ESG
accountability calculations. By providing accurate and consistent
sentiment scores, the model enhances transparency and comparability
across different periods and among various organizations. This is
important for stakeholders, including investors and regulators, who rely
on clear and credible ESG reporting. The model's ability
to identify key periods and provide objective criteria for performance
measurement allows for more strategic planning and targeted
interventions.

The findings also highlight the potential for integrating the Rasch
model's output with other ESG metrics and qualitative
data already used on the study area \cite{sokolov2020building,aue2022predicting, atanassova2024criticalminds}. This integration can provide
a comprehensive view of ESG performance, supporting a balanced scorecard
approach.

\section{Conclusion}

This exploratory study contributes to ESG research by introducing the
use of the Rasch model, a method from IRT, to analyze ESG sentiment
data. The study's findings highlight the
model's advantages in providing precise, reliable, and
consistent measurements compared to traditional approaches. By
effectively handling variability and offering deeper insights into the
latent traits of ESG sentiment, the Rasch model enables a more nuanced
and accurate assessment of ESG performance. This methodological
advancement enhances the accuracy of ESG accountability measures,
supports more informed decision-making processes, and strengthens
stakeholder trust through clearer and more credible reporting.

Future studies could further explore and refine the application of the
Rasch model in ESG sentiment analysis. One potential area for
advancement is the integration of more diverse data sources, such as
social media sentiment, stakeholder surveys, and financial performance
metrics, to enrich the analysis and provide a more comprehensive view of
ESG performance. Additionally, longitudinal studies could investigate
the stability and predictive power of ESG sentiment measured by the
Rasch model over time, assessing how these sentiments correlate with
actual ESG outcomes and broader market or societal trends.

\bibliographystyle{unsrt}  
\bibliography{references}  %%% Uncomment to use the external .bib file (using bibtex).

\end{document}